\documentclass[11pt]{article}

\usepackage[final]{acl}

\usepackage{times}
\usepackage{latexsym}

\usepackage[T1]{fontenc}
\usepackage[utf8]{inputenc}

\usepackage{microtype}

\usepackage{inconsolata}

\usepackage{xurl}

\usepackage{graphicx}
\usepackage{booktabs}
\usepackage{multirow}
\usepackage{xcolor}

\usepackage{amsmath}
\usepackage{amssymb}
\usepackage{pifont}

\usepackage{algorithm}
\usepackage{algorithmicx}
\usepackage{algpseudocode}

\newcommand{\cmark}{\ding{51}}%
\newcommand{\xmark}{\ding{55}}%

\newcommand{\gc}{\color{gray} \cmark}
\newcommand{\gx}{\color{gray} \xmark}

\newcommand{\redc}{\color{red} \cmark}
\newcommand{\redx}{\color{red} \xmark}

\title{HandReader: Advanced Techniques for Efficient Fingerspelling Recognition}

\author{
  Pavel Korotaev\textsuperscript{1} \\
  \texttt{pkorotaevdm@gmail.com}
  \And
  Petr Surovtsev\textsuperscript{1} \\
  \texttt{pysurovtsev@daimld.tech}
  \And
  Alexander Kapitanov\textsuperscript{1} \\
  \texttt{aakapitanov@daimld.tech} \\
  \AND
  Karina Kvanchiani\textsuperscript{1} \\
  \texttt{karinakvanciani@gmail.com}
  \And
  Alexander Nagaev\textsuperscript{1} \\
  \texttt{aonagaev@daimld.tech}
  \AND
\textsuperscript{1}\textnormal{Department of Artificial Intelligence and Machine Learning Development, DAIMLD}
}

\begin{document}
\maketitle

\begin{abstract}
Fingerspelling is a significant component of Sign Language (SL), allowing the interpretation of proper names, characterized by fast hand movements during signing. Although previous works on fingerspelling recognition have focused on processing the temporal dimension of videos, there remains room for improving the accuracy of these approaches. This paper introduces HandReader, a group of three architectures designed to address the fingerspelling recognition task. HandReader$_{RGB}$ employs the novel Temporal Shift-Adaptive Module (TSAM) to process RGB features from videos of varying lengths while preserving important sequential information. HandReader$_{KP}$ is built on the proposed Temporal Pose Encoder (TPE) operating on keypoints as tensors. Such keypoint composition in a batch allows the encoder to pass them through 2D and 3D convolution layers, utilizing temporal and spatial information and accumulating keypoint coordinates. We also introduce HandReader$_{RGB+KP}$ -- architecture with a joint encoder to benefit from RGB and keypoint modalities. Each HandReader model possesses distinct advantages: on the ChicagoFSWild and ChicagoFSWild+ datasets, HandReader models achieve the best results among methods trained on a single training set, including a new state-of-the-art result on ChicagoFSWild+.
\end{abstract}

\section{Introduction}
\label{sec:intro}

\begin{figure*}[t]
  \centering
  \includegraphics[width=1.0\linewidth]{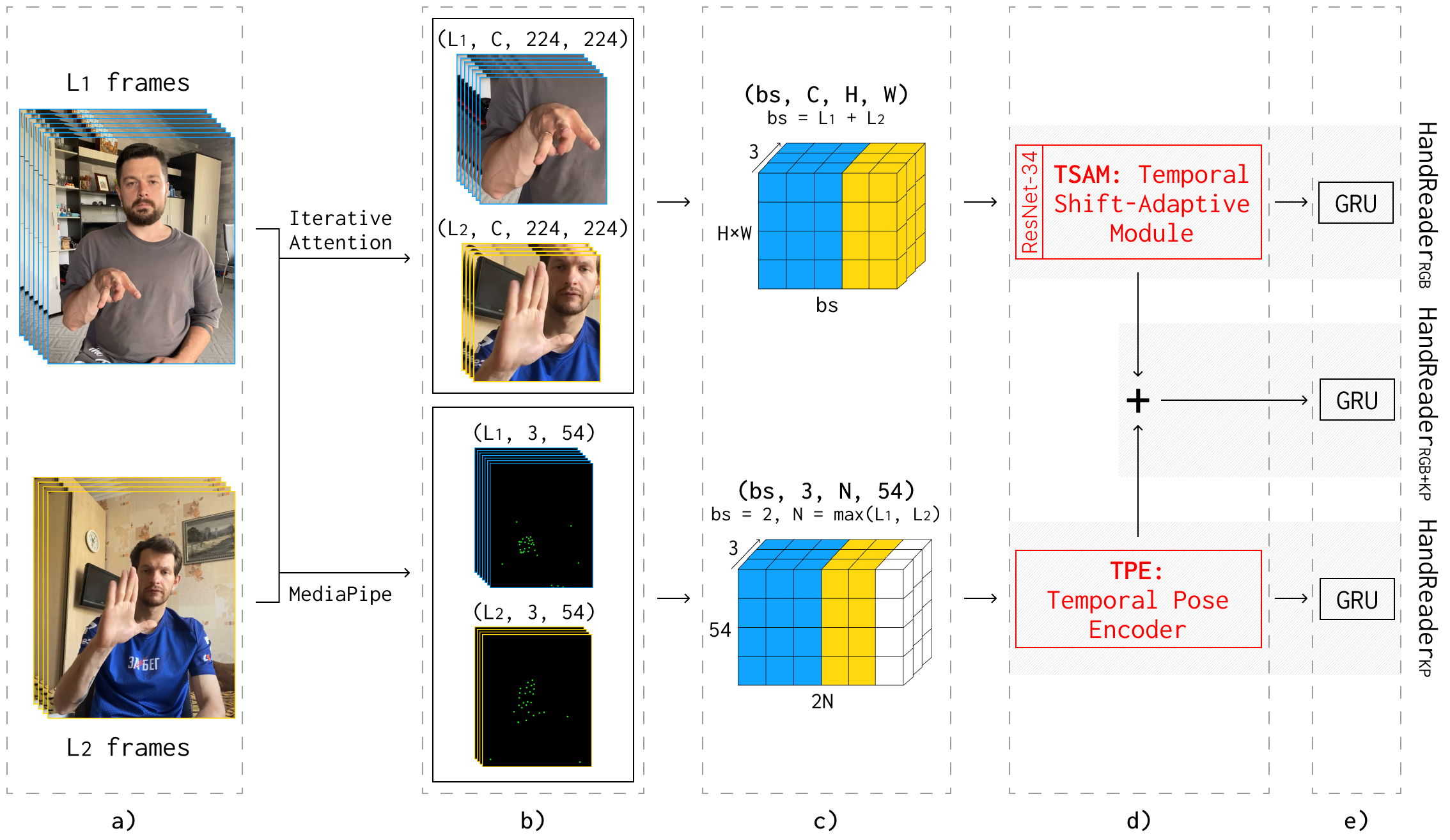}
  \caption{Three HandReader architectures. a) The videos with different lengths b) are processed to get crops and keypoints for HandReader$_{RGB}$ and HandReader$_{KP}$, respectively, c) form the batches, and d) pass through the encoders. Further, e) the GRU decoder is applied to the encoder outputs. The final output of the HandReader$_{RGB+KP}$ is obtained by summing the outputs of TSAM and TPE. Red-colored encoders are the paper's contribution.}
  \label{fig: overall}
\end{figure*}

Sign Language (SL) facilitates communication among more than 430 million deaf and hard-of-hearing individuals worldwide\footnote{\url{https://www.who.int/news-room/fact-sheets/detail/deafness-and-hearing-loss}}. Due to the limited quantity of SL interpreters, an automatic SL Recognition (SLR) system can bridge the communication barrier between deaf individuals and people without SL knowledge \citep{barrier1, barrier2, barrier3}. Such systems find application in schools, hospitals, and other social environments.

Fingerspelling is an integral component of any sign language, used for parts of speech, such as prepositions, conjunctions, and interjections, proper names, and instances without specific signs, such as addresses, abbreviations, and surnames. For example, to sign the metro station's name, the SL user should sign it letter by letter, using the corresponding signs for each letter from the SL alphabet. Therefore, fingerspelling is characterized by rapid and swift movements, complicating its recognition by blurry video frames. The motivation to tackle the fingerspelling recognition task is supported by interest from major companies, such as Google, which organized a Kaggle competition \citep{kaggle} to address it.

Many prior approaches to fingerspelling recognition are based on neural networks capable of processing the temporal component. To convert a video which consists of a frame sequence $I = \{I_1, I_2, ..., I_n\}$ to a letter sequence $W = \{W_1, W_2, ..., W_k\}$, encoder-decoder architectures are typically utilized \citep{posenet_5, kumwilaisak2022american_4, detect_as_backbone_2}. However, fingerspelling recognition, similar to Automatic Speech Recognition (ASR) and Optical Character Recognition (OCR), is complicated by the lack of alignment between $W$ and $I$, as multiple frames can interpret a single letter. The Connectionist Temporal Classification (CTC) loss \citep{CTC} function is widely used in ASR and OCR to address this challenge by facilitating the learning of alignment between input frames and their corresponding output sequences, without the need for explicitly annotated segmentations.
So, there is an option to simplify the task and enhance methods' performance by employing not only RGB frames but also such modalities as depth maps, keypoints, or combinations of video frames and keypoints \citep{depth_map_7, iter_attn_shi_1, posenet_5, regb_kp_6}.

Despite multiple attempts to enhance fingerspelling recognition, the results of prior methods \citep{depth_map_7, iter_attn_shi_1, posenet_5, regb_kp_6, detect_as_backbone_2, pannattee2024american_3, kumwilaisak2022american_4} on two American fingerspelling datasets -- ChicagoFSWild \citep{detect_as_backbone_2} and ChicagoFSWild+ \citep{iter_attn_shi_1} highlight the need for improved accuracy to ensure the applicability of existing approaches in real-world scenarios. Our work aims to enhance letter recognition accuracy while keeping the architectures lightweight.

We aimed to develop an efficient approach for fingerspelling recognition.
The contributions of this paper are summarized as follows:
\begin{itemize}
\item We introduce the Temporal Shift-Adaptive Module (TSAM), extending the original Temporal Shift Module (TSM) \citep{TSM} to effectively process RGB image features from videos of varying lengths without trimming or padding.
\item The Temporal Pose Encoder (TPE) to operate keypoints as three-channel tensors, capturing spatial and temporal information.
\item HandReader, a group of three architectures (see Figure~\ref{fig: overall}) that achieved a state-of-the-art result on the ChicagoFSWild+ test set and the best result on ChicagoFSWild among models trained solely on its training set.
\end{itemize}
We release the pre-trained models and code\footnote{\url{https://github.com/ai-forever/HandReader}}.

\section{Related Work}
\label{sec:related}
This overview excludes isolated Sign Language Recognition and Continuous Sign Language Recognition (CSLR) research, as methods for these tasks operate at the sign level: CSLR is further complicated by the need to recognize the independent relationships between signs and letters within words. In contrast, fingerspelling recognition presents a more straightforward one-to-one correspondence.

Recently, the fingerspelling recognition task has been explored using various data modalities, including RGB imagery \citep{iter_attn_shi_1, detect_as_backbone_2, pannattee2024american_3, kumwilaisak2022american_4}, keypoints \citep{posenet_5}, and depth maps \citep{depth_map_7}. Moreover, \citet{regb_kp_6} utilized data obtained by combining different modalities. Researchers are motivated to focus on the RGB modality because a fingerspelling recognition system must apply in real-world conditions and resist external changes. Meanwhile, acquiring data from other modalities requires additional equipment, such as extra cameras like Kinect for depth maps or an extra data processing stage for keypoints.

\begin{figure*}[t]
  \centering
  \includegraphics[width=1.0\linewidth]{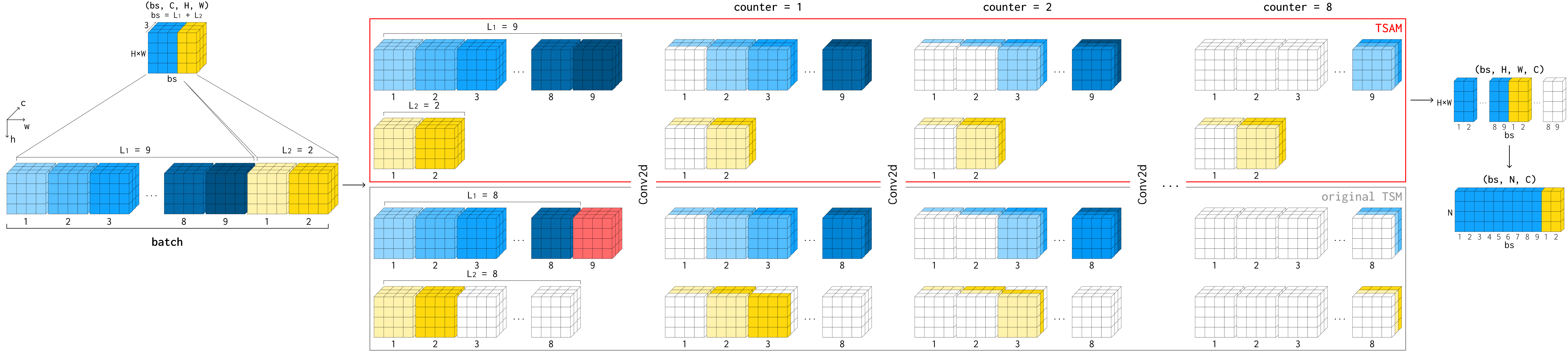}
  \caption{Comparison of the TSAM architecture (outlined in red at the top) with the original TSM (outlined in gray at the bottom). The input batch includes two videos of different lengths, with specific frames highlighted in blue and yellow. While the TSM model requires fixed-length inputs utilizing padding (shown in white) or trimming frames (shown in red) when necessary, the TSAM model employs individual shift counters to manage variable-length sequences without additional padding or trimming. The shift counter displayed at the top of the TSAM block shows how the shifts are activated individually for each video, incrementing the shift counter until it reaches the length of that specific video.}
  \label{fig: tsm}
\end{figure*}

Typical videos for the fingerspelling task contain irrelevant information, such as backgrounds and foreign objects. Therefore, only RGB data requires the model to identify the most relevant areas in the frames, specifically the hands, before recognizing a sign sequence. To simplify locating the signing hand, \citet{detect_as_backbone_2} developed a hand detection method using Faster R-CNN, then trained a CNN-LSTM model for the recognition task. In their subsequent work \citep{iter_attn_shi_1}, they replaced the explicit hand detection method with an iterative attention mechanism. The authors of \citet{pannattee2024american_3} and \citet{kumwilaisak2022american_4} employed representation learning to improve the model's ability to distinguish different signs by alignment from CTC and incorporated the method from \citet{iter_attn_shi_1} as a preprocessing stage. \citet{pannattee2024american_3} acquired state-of-the-art results on ChicagoFSWild and ChicagoFSWild+.

Since most RGB modality methods are centered around the hand detection problem, they are limited to exploring different architectures or decoding methods on the final results. In contrast, the keypoint modality avoids this issue because it includes relevant information about hand placement. \citet{posenet_5} trained a Transformer encoder-decoder architecture with a new decoding approach, advancing performance on ChicagoFSWild and ChicagoFSWild+ test sets. Keypoints are usually acquired using MediaPipe \citep{mediapipe} or another neural network trained for pose estimation. \citet{regb_kp_6} combined RGB and keypoint modalities employing linear layers to project the 2D coordinates into 3D space. After extracting all features, they concatenated them and used a decoder based on an attention mechanism. Despite the promise shown by the methods above, recognition accuracy still has potential for improvement. Therefore, our primary objective is to enhance the methodologies employed without compromising the model's weight and inference speed.

\noindent\textbf{Multi-Stream Sign Language Models.} Combining RGB and keypoint streams has also been explored beyond fingerspelling. \citet{stmc} exploited multiple visual cues, including hands, face, and full-frame appearance, for continuous sign language recognition and translation, while \citet{twostream} modeled raw videos and keypoint sequences with a two-stream network for the same tasks. More recently, self-supervised multi-stream approaches, such as SignMusketeers \citep{signmusketeers} and SHuBERT \citep{shubert}, learn sign language representations from separate face, hand, and body streams. These methods operate at the sign or sentence level and typically involve large-scale pre-training. In contrast, our work addresses the letter-level fingerspelling task, and its contribution lies at the encoder level: TSAM for variable-length RGB processing and TPE for tensor-based keypoint encoding, combined through a deliberately lightweight fusion.

\noindent\textbf{RGB Feature Extraction in the Fingerspelling Task.} There are two main ways to extract features in the fingerspelling task. (1) Spatial features can be received by applying a 2D CNN to each frame, then extracting temporal features using 1D convolutions or RNNs. (2) Spatio-temporal features can be obtained with 3D convolution. However, processing frames as a sequence of independent frames lacks temporal information. The TSM \citep{TSM} was developed to address this issue by employing 2D convolutions and shifting information along the temporal dimension. However, TSM processes fixed-size videos, which leads to the loss of valuable information in long videos or negatively affects training time and memory allocation due to the padding in shorter videos. Therefore, this paper presents a modified version of the original TSM to handle videos of varying lengths.

\begin{figure*}[t]
  \centering
  \includegraphics[width=1.0\linewidth]{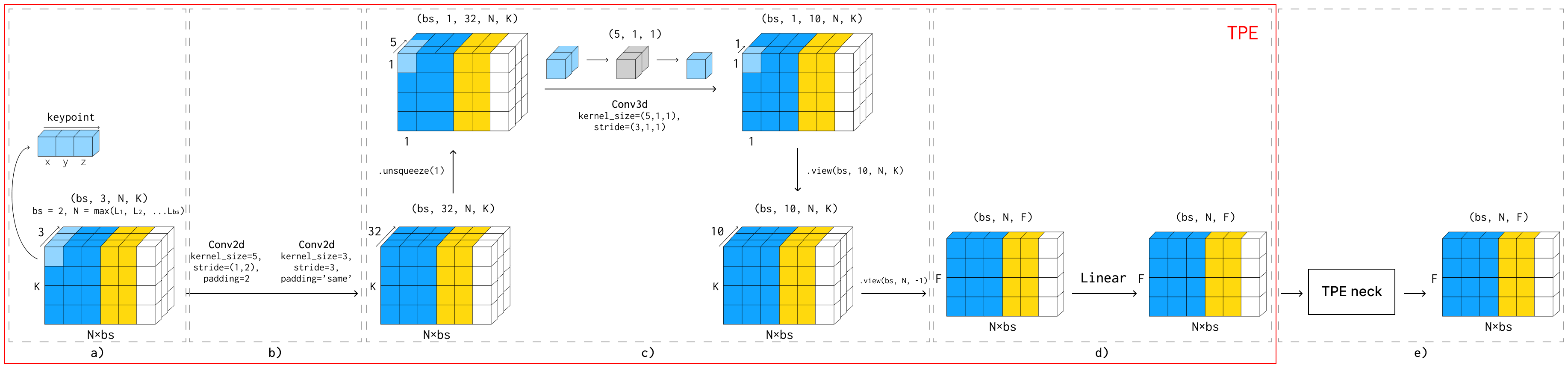}
  \caption{Demonstration of the Temporal Pose Encoder operating on a single batch. a) The encoder receives an input tensor containing $N$ frames and $K$ keypoints. All videos are padded to match the length of the most extended video clip in the batch, allowing them to be stacked. b) 2D convolutions process multiple keypoints across several frames, extracting temporal features. c) Subsequently, a 3D convolution in the form of a ``tube'' aggregates the extracted motion information for each body part into final feature representations $F$. d) After applying TPE, features are processed with linear layer, e) followed by MLP with convolution modules from \citet{Conformer}.}
  \label{fig: kp_arch}
\end{figure*}

\section{Approach}
\label{sec:approach}

We present a family of three architectures, collectively referred to as HandReader, designed for sign sequence recognition across multiple modalities. The models differ only in the encoder stage (see Figure~\ref{fig: overall}). Thus, HandReader$_{RGB}$ solely utilizes RGB features, HandReader$_{KP}$ relies on keypoints, and HandReader$_{RGB+KP}$ combines both. Each was trained directly using CTC loss, eliminating the need for frame-level letter alignment. The CTC loss accounts for all possible alignments between hand shapes and sign language letters, eliminating the need for explicit alignment supervision. More formally:

\begin{equation}
L_{CTC} = - \log \sum_{A \in \mathcal{A}_{X,Y}} \prod^{T}_{t=1} p_t(a_t \mid X),
\end{equation}
where $\mathcal{A}_{X,Y}$ represents the set of valid alignments between the model input $X$ and the target sequence $Y$, $A = (a_1, \dots, a_T)$ is one such alignment, and $p_t(a_t \mid X)$ denotes the probability of symbol $a_t$ occurring at time step $t$. Each $a_t$ is either a letter of the alphabet or the special blank symbol ``\_'', introduced by CTC for time steps that emit no letter and to separate repeated letters; blanks and repetitions are removed from an alignment to obtain the final prediction.

A two-layer bidirectional Gated Recurrent Unit (GRU) \citep{GRU} was chosen as a decoder to generate a sequence of characters, enhancing the model's capability to capture temporal dynamics. We apply a linear layer with SoftMax to the GRU output to generate posterior probabilities for each predicted character at every frame. Encoders for each architecture are described below.

\subsection[HandReader RGB]{HandReader$_{\textbf{RGB}}$}
\label{HandReader-RGB}

The original TSM is a video understanding approach that enables temporal modeling in 2D convolutional networks by shifting portions of channel features along the temporal dimension. However, TSM is constrained to processing a fixed number of frames. Therefore, videos that exceed the frame limit require trimming, while shorter videos need additional empty frames (i.e., padding). Furthermore, the number of shifts is determined by the fixed frame limit. As a result, videos with extra empty frames lose critical sequential information during processing (see Figure~\ref{fig: tsm}). We modified the TSM to manage a dynamic number of input frames, avoiding padding and enabling frame-by-frame processing for each video in the batch. Besides, a shift counter is introduced to preserve important sequential information, bypassing shifts on short videos.

\noindent\textbf{Temporal Shift-Adaptive Module.} The TSAM was designed to effectively process videos of various lengths. Figure~\ref{fig: tsm} illustrates the distinction between TSAM and TSM operations. The TSAM is incorporated into the ResNet-34 model and operates an input tensor with dimension $(bs, C, H, W)$, where $bs$ indicates batch size, equal to the sum of the video lengths, $C$ is the number of channels, $H$ and $W$ are the height and width of the frames, respectively. The TSAM also requires an auxiliary list of video lengths, denoted as $L = [L_{S_1}, L_{S_2}, \dots, L_{S_n}]$, where $S_i$ signifies $i$th video in the batch, and $bs = \sum_i L_{S_i}$. We iterate through $L$, extracting $L_{S_i}$ frames from the batch to obtain $S_i$ video sequence and apply shifts individually, as detailed in Algorithm~\ref{alg: tsam}. Such a method reduces VRAM usage by up to $50\%$ during batch creation, thereby facilitating the processing of longer videos.

The number of shifts is determined by the number of convolutional blocks within the backbone architecture, leading to overwrite temporal features similar to the original TSM. To prevent this loss of information, we introduce a shift counter that stops temporal shifts at the final frame, i.e., no shifting is applied if a video's length is smaller than the current shift count. Such a method ensures that each video in a batch is treated individually, resulting in more accurate feature extraction.

Each sequence of the received tensor with dimension $(bs, H, W, F)$, where $F$ represents the number of features, was subsequently padded with zeros to align with the maximum sequence length within the batch. All features are aggregated using 1D convolution, and the resulting tensor is transformed into the format $(bs, N, F)$, where $F$ is the flattened features across all channels, and $N$ is calculated as $H * W$.

\begin{algorithm}
\caption{Applying temporal shifts in TSAM.}
\label{alg: tsam}
\begin{algorithmic}[1]
\Require Input tensor $S$ of size $(bs, C, H, W)$, list of sequence lengths $L = [L_{S_1}, L_{S_2}, \dots, L_{S_n}]$
\Ensure Output tensor with temporal shifts applied

\For{each sequence $S_i$ in batch from input tensor}
    \State $counter \gets 0$
    \For{each convolutional block in backbone}
        \If{$counter < L_{S_i}$}
            \State Apply temporal shift for sequence $S_i$
            \State $counter \gets counter + 1$
        \Else
            \State Skip temporal shift for sequence $S_i$
        \EndIf
    \EndFor
    \State Apply 1D Convolution for $S_i$
\EndFor
\end{algorithmic}

\end{algorithm}

\subsection[HandReader KP]{HandReader$_{\textbf{KP}}$}
We utilized MediaPipe \citep{mediapipe} to extract body keypoints from every frame. Face keypoints were excluded as stated in Section~\ref{ablation_kp}; we used 21 keypoints for each hand and 12 body pose keypoints (6 per arm), resulting in 54 $(x, y, z)$ coordinates per frame. The proposed feature encoder is built on a new way to structure keypoints explicitly developed for it, which organizes keypoints within tensors. Each video is treated as a tensor with dimension $(3, N, 54)$, where 54 $(x, y, z)$ vectors are stacked into matrices for each of $N$ video frames. Such a configuration enables the spatial representation of temporal data, allowing the application of processing techniques typically used for standard images. We passed the received tensor to the feature extractor described below.

\noindent\textbf{Temporal Pose Encoder.} We present a Temporal Pose Encoder based on the convolution layers. TPE operates an input tensor of dimension $(bs, 3, N, K)$, where $bs$ is the batch size, indicating the number of videos in the batch (in contrast to TSAM, where $bs$ denotes the total number of frames across the videos in the batch), $N$ denotes the maximum length of videos in the batch, while shorter videos were padded, and $K$ -- is the number of keypoint coordinates. The tensor is processed by two 2D convolution layers, followed by a 3D one (see Figure~\ref{fig: kp_arch}). 2D convolutions serve to extract both temporal and spatial features, while transforming keypoint coordinates, enhancing the model's understanding of dynamic finger movements. At the same time, 3D convolution tries to accumulate coordinate information within a single video with the kernel-tube size of $(5, 1, 1)$, treating each coordinate independently. In this context, the hyperparameter of 5 defines the size of the receptive window along the coordinate dimension, which captures local patterns across five neighboring coordinate features. We insert a dummy dimension in the second position, enabling 3D convolution to treat the original channels as temporal dimensions. The process can be represented as follows: $(bs, 32, N, K)$ is unsqueezed to $(bs, 1, 32, N, K)$ and processed by 3D convolution to $(bs, 1, 10, N, K)$. A linear layer is then applied to project the features of the keypoints $(bs, N, F)$, where $F$ is the final encoder's features. We additionally use multi-layer perceptron (MLP) with convolution modules from \citet{Conformer} as a neck between the encoder and the GRU decoder.

\subsection[HandReader RGB+KP]{HandReader$_{\textbf{RGB+KP}}$}
We combine features from the HandReader$_{RGB}$ and HandReader$_{KP}$ encoders to take advantage of both modalities. Two output tensors of features from TSAM and TPE are summed and passed to the GRU (see Figure~\ref{fig: overall}).

\section{Experiments}
\label{sec:experiments}
This section outlines the experimental setup and data preprocessing procedures used to assess the proposed HandReader architectures and compare them with other approaches. Table~\ref{tabl:datasets} shows the open datasets used in this work.

\subsection{Experimental Setup \label{ExpSetup}}
All three architectures were trained from scratch using an AdamW optimizer \citep{adamw} with an initial learning rate of 0.0001. The MultiStepLR modifies the learning rate with a gamma of 0.1 using milestones at 20 and 40 for the HandReader$_{RGB}$ and HandReader$_{RGB+KP}$ models and 25 and 40 for the HandReader$_{KP}$. We trained the HandReader$_{RGB}$ for 60 epochs, while the other two models were trained for 100 epochs each. A single Nvidia H100 with 80GB RAM was utilized for training.

\begin{table}[tb]
\begin{center}
\resizebox{\columnwidth}{!}{
\begin{tabular}{lcccccc}
\hline
\textbf{Dataset} & \textbf{Lang} & \textbf{Sequences} & \textbf{Chars} & \textbf{Hours} & \textbf{Signers} &\textbf{Source}\\
\hline\hline
\text{ChicagoFSWild} & \text{ASL} & \text{7K} & \text{38K} & \text{2} & \text{160} & \text{Web} \\
\text{ChicagoFSWild+} & \text{ASL} & \textbf{55K} & \text{0.3M}& \text{14} & \textbf{260} & \text{Web}\\
\hline
\end{tabular}
}
\caption{Open fingerspelling recognition datasets used in this work.}
\label{tabl:datasets}
\end{center}
\end{table}

\subsection{Data Preprocessing}
\label{DataPreproc}
The ChicagoFSWild and ChicagoFSWild+ datasets are commonly used to evaluate approaches to the fingerspelling task.

The RGB frames for each dataset were cropped using the model described in \citet{iter_attn_shi_1}, augmented with horizontal flipping and random rotation (see parameters in Table~\ref{tab:augs}), then resized to 224 and normalized. Keypoints were extracted by MediaPipe from raw frames rather than cropped images. We utilized only the hand keypoints (21 per hand) and 12 body pose keypoints (6 per arm), resulting in 54 keypoints. We replaced the NaN values with zeros, normalized the rest, and applied augmentations specified in Table~\ref{tab:augs}. Only resampling, horizontal flipping, and random rotation augmentations are applied when combining two modalities in HandReader$_{RGB+KP}$.

\begin{table}[tb]
\begin{center}
\resizebox{\columnwidth}{!}{
\begin{tabular}{l c c c}
 \hline
 \textbf{Augmentation} & \textbf{Parameters} & \textbf{Probability } & \textbf{Modality} \\ [0.5ex]
 \hline\hline
 Resample & rate=(0.5, 1.5) & 0.8 & KP, RGB+KP\\ \hline
 SpatialAffine & \begin{tabular}[c]{@{}c@{}}scale=(0.8, 1.2), shear=(-0.15, 0.15), \\ shift=(-0.1, 0.1), degree=(-30, 30)\end{tabular} & 0.75  & KP \\ \hline
 TemporalMask & size=(0.2, 0.4), value=0.0 & 0.5 & KP\\ \hline
 SpatialMask & \begin{tabular}[c]{@{}c@{}}size=(0.05, 0.1), \\ value=0.0, mode="relative"\end{tabular} & 0.5 & KP \\ \hline
 HorizontalFlip & --- & 0.5 & RGB, RGB+KP, KP \\ \hline
 Rotation & angle=(-10, 10) & 0.5 & RGB, RGB+KP \\
\hline
\end{tabular}}
\caption{Augmentations used during training.}
\label{tab:augs}
\end{center}
\end{table}

\subsection{Results \label{Results}}
We assessed HandReader architectures using the letter accuracy metric: $1 - \frac{S+D+I}{N}$, where $S$, $D$, and $I$ denote substitutions, deletions, and insertions, respectively, $N$ is the length of ground-truth sequence. Metric values for other methods were obtained from the original papers for comparison purposes. Each HandReader version outperformed all prior methods trained on a single training set on both benchmarks, and HandReader$_{RGB+KP}$ achieved a new state-of-the-art result on ChicagoFSWild+ (see Table~\ref{tab:result_chicago}). Note that Conformers (CFW+CFW+) \citep{shi2023americansignlanguageprocessing} reaches a higher score on ChicagoFSWild; however, it was trained on the union of the ChicagoFSWild and ChicagoFSWild+ training sets, whereas each HandReader model was trained only on the training set of the corresponding benchmark.
We also present prediction comparison between PoseBased Transformer \citep{posenet_5} and our models in Table~\ref{tabl:posenet_comp_preds_table}. For decoding, all models employed the beam search strategy with a beam width of 5, which explores multiple candidate sequences at each step and selects the most probable transcription.

Table~\ref{tabl:comparison_results2} provides qualitative examples of predictions produced by the three HandReader variants on the ChicagoFSWild dataset. We observe that the keypoint-based model (\textbf{KP}) tends to generate shorter and less stable outputs, frequently dropping characters (e.g., ``youtube'' $\rightarrow$ ``outube'') or producing heavily distorted strings. The RGB-based model (\textbf{RGB}) is generally more consistent in capturing the global temporal structure, but still suffers from confusions between visually similar finger configurations, resulting in character-level substitutions (e.g., ``organic'' $\rightarrow$ ``orgadic''). The multimodal model (\textbf{RGB+KP}) demonstrates the most robust behavior across the examples, often correcting errors present in the unimodal variants, which indicates that RGB and keypoint cues provide complementary information.

To further analyze the error patterns, Table~\ref{tab:sdi_chicagofswild} reports the SDI (Substitution, Deletion, Insertion) breakdown for each model under greedy and beam-search decoding. Across all models, \textbf{deletions} constitute the dominant error type, highlighting the difficulty of maintaining correct character continuity in fast fingerspelling sequences. Beam search consistently reduces the number of deletions for all variants (e.g., from 677 to 628 for RGB, and from 857 to 789 for KP), at the cost of a small increase in substitutions and insertions. This trade-off suggests that beam search improves sequence completeness but may introduce additional character-level confusions. Overall, the \textbf{RGB+KP} model achieves the best balance between the three error types, reinforcing the benefit of multimodal fusion for fingerspelling recognition.

\begin{table*}[t]
\begin{center}
\scalebox{0.85}{
\begin{tabular}{l c c c}
 \hline
 \textbf{Model} & \textbf{ChicagoFSWild} & \textbf{ChicagoFSWild+} & \textbf{Modality} \\ [0.5ex]
 \hline\hline
 Hand Det + CNN + RNN \citep{detect_as_backbone_2} & 41.9 & 41.2 & RGB \\
 Iterative Attention \citep{iter_attn_shi_1} & 45.1 & 46.7  & RGB \\
 3d hand pose + RGB \citep{regb_kp_6} & 47.9 & -- & RGB + KP\\
 SiameseNet+Iterative model \citep{kumwilaisak2022american_4} & 49.6 & -- & RGB \\
 Iter train + SupCon \citep{pannattee2024american_3} & 64.9 & 71.7*  & RGB \\
 Pose-based transformer \citep{posenet_5} & 66.3 & 71.1  & KP \\
 Conformers (CFW) \citep{shi2023americansignlanguageprocessing} & 68.4 & 60.2 & RGB \\
 Conformers (CFW+CFW+) \citep{shi2023americansignlanguageprocessing} & 77.5* & 73.0* & RGB \\
 \textbf{HandReader$_{\textbf{KP}}$} & \textbf{69.3} & \textbf{72.4} & KP \\
\textbf{HandReader$_{\textbf{RGB}}$} & \textbf{72.0} & \textbf{73.8} & RGB \\
\textbf{HandReader$_{\textbf{RGB+KP}}$} & \textbf{72.9}** & \textbf{75.6} & RGB + KP\\
\hline
\end{tabular}}
\caption{Results comparison on ChicagoFSWild and ChicagoFSWild+ test sets. * indicates the model was trained on both ChicagoFSWild and ChicagoFSWild+ training sets. ** means two convolution modules were used for HandReader$_{KP}$ encoder, while only one was employed for other metrics.}
\label{tab:result_chicago}
\end{center}
\end{table*}

\begin{table}[tb]
\begin{center}
\resizebox{\columnwidth}{!}{
\begin{tabular}{lcccc}
\hline
\textbf{PoseBased \citep{posenet_5}*} & \textbf{KP} & \textbf{RGB+KP} & \textbf{RGB} & \textbf{Reference}\\
\hline\hline
macah & mach & \text{mach} & mach & mach  \\
meach & meah & beack & fealc & beach \\
hospital  & hospital & hospital & hospital & hospital\\
- & outube & iutube & iutube & youtube \\
- & organic & organic & orgadic & organic \\
\hline
\end{tabular}
}
\caption{Results comparison of predictions between previous SotA solution and our models, predictions for PoseBased model were taken from the original paper.}
\label{tabl:posenet_comp_preds_table}
\end{center}
\end{table}

\begin{table}[t]
\begin{center}
\resizebox{\columnwidth}{!}{
\setlength{\tabcolsep}{12pt}
\begin{tabular}{lccc}
\hline
\textbf{Gt} & \textbf{KP} & \textbf{RGB} & \textbf{RGB+KP} \\
\text{youtube} & \text{outube} & \text{iutube} & \text{iutube}\\
\text{issue} & \text{isue} & \text{isue} & \text{isue} \\
\text{organic} & \text{or ganic} & \text{orgadic} & \text{organic} \\
\text{pharmacology} & \text{opharmacoint} & \text{opharoacolohy} & \text{opharmacolgi} \\
\text{pharmacology} & \text{harmcogiy} & \text{haroscothy} & \text{harmacotgy} \\
\text{physical} & \text{phisical} & \text{phylsical} & \text{phistical} \\
\text{audio} & \text{uio} & \text{ardio} & \text{atudio} \\
\text{biochemistry} & \text{biochemitry} & \text{biockemistry} & \text{ biochemiry}\\
\hline
\end{tabular}
}
\caption{Result of comparison of predictions between different models on ChicagoFSWild dataset.}
\label{tabl:comparison_results2}
\end{center}
\end{table}

\begin{table}[t]
\centering
\resizebox{\columnwidth}{!}{
\setlength{\tabcolsep}{12pt}
\begin{tabular}{lccc}
\hline
\textbf{Model} & \textbf{D} & \textbf{S} & \textbf{I} \\
\hline
RGB      & 677 / 628 & 462 / 478 & 122 / 130 \\
KP       & 857 / 789 & 382 / 406 & 148 / 161 \\
RGB+KP   & 683 / 621 & 386 / 409 & 160 / 164 \\
\hline
\end{tabular}
}
\caption{SDI breakdown on ChicagoFSWild. Each cell shows Greedy / Beam decoding.}
\label{tab:sdi_chicagofswild}
\end{table}

\section{Ablation}
\label{sec:ablation}
This section provides an ablation study to demonstrate the impact of each component on the HandReader's overall performance. Note that HandReader$_{RGB+KP}$ was excluded from the study since it relies solely on the heuristics of the other two architectures. We maintained a consistent experimental setup described in Section~\ref{sec:experiments} and changed only one component at a time to analyze its influence. The ChicagoFSWild dataset was utilized in all experiments.

\subsection[HandReader RGB]{HandReader$_{\textbf{RGB}}$}
\label{ablation_rgb}
\textbf{Encoder.} Table~\ref{tab:ablation_rgb_1} indicates that integrating TSM into the baseline model improves the performance metric by 1.69. Furthermore, replacing TSM with TSAM leads to an additional 1.4 improvement. To isolate the effect of the count shift mechanism, we implemented a variant of TSAM without the timely stopping of shifts, which led to a 1.49 performance degradation.

\noindent\textbf{Decoder.} We experimented with the LSTM \citep{LSTM} decoder instead of the GRU \citep{GRU} one, which resulted in a difference of 1.69. Replacing RNN with a linear layer reduces the metric by 11.0. Such observations underlined that high-level features of each frame cannot be used as independent features for final predictions.

\begin{table*}[tb]
\begin{center}
\scalebox{0.8}{
\begin{tabular}{l c c c c c c c}
\hline
\textbf{Experiment} & \textbf{\begin{tabular}[c]{@{}c@{}} Packed sequence\end{tabular}} & \textbf{RNN} & \textbf{\begin{tabular}[c]{@{}c@{}}Count shift\end{tabular}} & \textbf{\begin{tabular}[c]{@{}c@{}}Shift type\end{tabular}} & \textbf{\begin{tabular}[c]{@{}c@{}}Image\\ Augs.\end{tabular}} & \textbf{BS} & \textbf{Letter accuracy} \\ \hline\hline
\textbf{best} & \textbf{\cmark} & \textbf{GRU} & \textbf{\cmark} & \color{gray}TSAM & \textbf{\cmark} & \textbf{4} & \textbf{72.0} \\ \hline\hline
\textbf{\begin{tabular}[c]{@{}c@{}}Packed sequence\end{tabular}} & \redx & \color{gray} GRU & \redx & \color{red}TSM & \gc & \color{gray} 4 & \(70.6_{\color{red}-1.4}\) \\ \hline
\multirow{2}{*}{\textbf{RNN}} & \gc & \color{red} LSTM & \gc  & \color{gray}TSAM & \gc & \color{gray} 4 & \(70.31_{\color{red}-1.69}\)\\
 & \gc & \redx & \gc  & \color{gray}TSAM & \gc & \color{gray} 4 & \(61.0_{\color{red}-11.0}\) \\ \hline
\textbf{\begin{tabular}[c]{@{}c@{}}Count shift \end{tabular}} & \gc & \color{gray} GRU & \redx  & \color{gray}TSAM & \gc & \color{gray} 4 & \(70.51_{\color{red}-1.49}\)\\ \hline
\textbf{\begin{tabular}[c]{@{}c@{}}Image Augs.\end{tabular}} & \gc & \color{gray} GRU & \gc  & \color{gray}TSAM & \redx & \color{gray} 4 & \(70.49_{\color{red}-1.51}\) \\ \hline
\multirow{2}{*}{\textbf{BS}} & \gc & \color{gray} GRU & \gc   & \color{gray}TSAM & \gc & \color{red} 2 & \(71.46_{\color{red}-0.54}\) \\
 & \gc & \color{gray} GRU & \gc  & \color{gray}TSAM & \gc & \color{red} 6 & \(70.1_{\color{red}-1.9}\) \\ \hline
 \textbf{\begin{tabular}[c]{@{}c@{}}Baseline*\end{tabular}} & \redx & \color{gray} GRU & \redx & \color{red}None & \gc & \color{gray} 4 & \(69.63_{\color{red}-2.37}\) \\ \hline
\end{tabular}}
\caption{Ablation study of HandReader$_{RGB}$ components. * denotes the baseline setup with a ResNet34 configuration (without TSM or TSAM shift operations), which already outperforms previous state-of-the-art solutions. Here, we define BS as batch size and RNN as the type of recurrent neural network used, which may include LSTM, GRU, or a linear layer without recurrence. ``Packed sequence'' refers to a modification of the TSM mechanism that enables processing each video independently during the shift operation, as described in Section~\ref{HandReader-RGB}.}
\label{tab:ablation_rgb_1}
\end{center}
\end{table*}

\begin{table*}[tb]
\begin{center}
\scalebox{0.8}{
\begin{tabular}{c ccc c c c c c}
\hline
\multirow{2}{*}{\textbf{Experiment}} & \multicolumn{3}{c}{\textbf{Keypoints}} & \multirow{2}{*}{\textbf{RNN}} & \multirow{2}{*}{\textbf{\begin{tabular}[c]{@{}c@{}}Conv. \\ modules\end{tabular}}} & \multirow{2}{*}{\textbf{Image Augs.}} & \multirow{2}{*}{\textbf{BS}} & \multirow{2}{*}{\textbf{Letter accuracy}} \\ \cline{2-4}
 & \textbf{hand} & \textbf{pose} & \textbf{face} &  &  &  &  & \\ \hline\hline
\textbf{best} & \textbf{\cmark} & \textbf{\cmark} & \textbf{\xmark} & \textbf{GRU} & \textbf{1} & \textbf{\cmark} & \textbf{4} & \textbf{69.31} \\ \hline\hline
\multirow{6}{*}{\textbf{Keypoints}} & \gc &\redx& \gx & \multirow{6}{*}{\color{gray}GRU} & \multirow{6}{*}{\color{gray}1} & \gc  & \multirow{6}{*}{\color{gray}4} & \(68.54_{\color{red}-0.77}\)\\
 & \gc & \gc & \redc &  &  & \gc  &  & \(69.08_{\color{red}-0.23}\)\\
 & \gc &\redx& \redc &  &  & \gc  &  & \(67.59_{\color{red}-1.72}\)\\
 &\redx& \gc & \redc &  &  & \gc  &  & \(10.02_{\color{red}-59.29}\)\\
 &\redx& \gc & \gx &  &  & \gc  &  & \(10.36_{\color{red}-58.95}\)\\
 &\redx&\redx& \redc &  &  & \gc  &  &
 \( 11.08_{\color{red}-58.23}\)\\ \hline
\multirow{2}{*}{\textbf{RNN}} & \gc & \gc & \gx &\color{red} LSTM & \multirow{2}{*}{\color{gray}1} & \gc  & \multirow{2}{*}{\color{gray}4} & \(67.79_{\color{red}-1.52}\)\\
 & \gc & \gc & \gx &\redx&  & \gc  &  & \(63.88_{\color{red}-5.43}\)\\ \hline
\multirow{2}{*}{\textbf{\begin{tabular}[c]{@{}c@{}}Conv. modules\end{tabular}}} & \gc & \gc & \gx & \multirow{2}{*}{\color{gray}GRU} &\color{red} 2 & \gc  & \multirow{2}{*}{\color{gray}4} & \(68.09_{\color{red}-1.22}\)\\
 & \gc & \gc & \gx &  &\color{red} 3 & \gc  &  & \(68.11_{\color{red}-1.2}\)\\ \hline
\textbf{Image Augs} & \gc & \gc & \gx & \color{gray}GRU & \color{gray}1 & \redx  & \color{gray}4 & \(62.43_{\color{red}-6.88}\)\\\hline
\multirow{3}{*}{\textbf{BS}} & \gc & \gc & \gx & \multirow{3}{*}{\color{gray}GRU} & \multirow{3}{*}{\color{gray}1} & \gc  &\color{red} 2 & \(69.04_{\color{red}-0.27}\)\\
 & \gc & \gc & \gx &  &  & \gc  &\color{red} 6 & \(69.26_{\color{red}-0.05}\)\\
 & \gc & \gc & \gx &  &  & \gc  &\color{red} 8 & \(68.45_{\color{red}-0.86}\)\\ \hline
\end{tabular}}
\caption{Ablation study of HandReader$_{KP}$ components. Here, ``Conv Modules'' denotes the number of convolutional modules from the Conformer architecture that are used inside the encoder. Other terms are defined in Table~\ref{tab:ablation_rgb_1}.}
\label{tab:ablation_kp_1}
\end{center}
\end{table*}

\noindent\textbf{Others.} Table~\ref{tab:ablation_rgb_1} highlights the significant impact of augmentations, as a model trained without rotations and horizontal flips reached a final metric of 70.49. Such influence is explained by the dataset containing signers with different dominant hands and videos where hands are misaligned due to camera angles or the signer's position. Additionally, we tested different batch sizes and showed that a bigger batch size is not always optimal. It is especially noticeable in ablation studies with RGB. This may happen because the batch contains videos of varying lengths, complicating the model's decoding process.

\subsection[HandReader KP]{HandReader$_{\textbf{KP}}$}
\label{ablation_kp}
\textbf{Keypoints.} We tested possible combinations of different parts of keypoints and provided results in Table~\ref{tab:ablation_kp_1}. The metric was reduced by adding face keypoints since sign language users can express the same phrases differently without verbal language skills. The model performs better with pose and hand keypoints than with just hand keypoints, as the pose coordinates carry information about the position of the whole hand, which may help the model better understand the hand's relative position throughout the video.

\noindent\textbf{Encoder and Decoder.} The same procedure of RNN impact assessment is provided in Table~\ref{tab:ablation_kp_1}. In contrast to Section~\ref{ablation_rgb}, the usage of the linear layer does not show a significant metric drop, as the HandReader$_{KP}$ encoder contains more modules for processing temporal dimensions. We studied the impact of the number of convolution modules utilized in the encoder. As the number of convolution blocks increases, the final metric declines.

\noindent\textbf{Others.} The augmentations have a crucial effect on model performance. We also examined various batch sizes: 4, 6, and 8. As the batch size increases, there are noticeable metric drops.

\subsection[HandReader RGB+KP]{HandReader$_{\textbf{RGB+KP}}$}
\label{ablation_kp_rgb}
\textbf{Feature reduction.}
We evaluated various feature reduction techniques, including simple summation, feature concatenation, element-wise multiplication, and weighted summation, obtaining 72.95, 72.95, 72.55, and 72.57 letter accuracy, respectively. We preferred the simple summation due to the smaller dimension of the sum output than concatenation one. The evaluation was conducted on the test set of the ChicagoFSWild dataset.

\section{Conclusion}
This paper presents HandReader, a suite of three distinct architectures that significantly advance the fingerspelling recognition task. The Temporal Shift-Adaptive Module and Temporal Pose Encoder were developed to capture RGB and keypoint features effectively. We achieved state-of-the-art performance on the ChicagoFSWild+ dataset and the best performance on ChicagoFSWild among models trained solely on its training set. The ablation study revealed that the proposed techniques and training pipelines are essential for achieving optimal results. The presented findings offer valuable insights for enhancing fingerspelling recognition systems.

% In the ACL layout, the Limitations section is unnumbered and placed
% after the conclusion, before the references. It does not count toward
% the page limit.
\section*{Limitations}
\label{sec:limitations}

\textbf{Architectures.} Although the HandReader$_{KP}$ and HandReader$_{RGB}$ effectively recognize fingerspelling, extra models are required to extract keypoints and hand region crops, respectively. Such necessities add computational overhead, decreasing inference speed and limiting the model's real-time usability, particularly on devices with limited processing power. There is an opportunity to replace an additional model with center cropping augmentation in the HandReader$_{RGB}$, provided the input focuses on the hands' region in the center of the frame. The HandReader$_{RGB+KP}$ is subject to the limits of other HandReader architectures, but it can utilize only the keypoint prediction model and obtain crops operating received points.

% ---- Bibliography ----
% The acl.sty style file already selects the acl_natbib bibliography style.
\bibliography{main}

\end{document}